%% file: main.tex
\documentclass[11pt]{article}

\usepackage[margin=0.9in]{geometry}
\usepackage{times}
\usepackage{graphicx}
\usepackage{amsmath,amssymb}
\usepackage{booktabs}
\usepackage{multirow}
\usepackage{hyperref}
\usepackage{microtype}
\usepackage{xcolor}
\usepackage{enumitem}
\usepackage{array}
\usepackage{subcaption}
\usepackage[numbers,sort&compress]{natbib}
 
\hypersetup{
  colorlinks=true,
  linkcolor=blue!60!black,
  citecolor=blue!60!black,
  urlcolor=blue!60!black
}

\title{
  \textbf{Probabilistic Dating of Historical Manuscripts via\\
  Evidential Deep Regression on Visual Script Features}
}
 
\author{
  Ranjith Chodavarapu\\
  Kent State University\\
  \texttt{[rchodava@kent.edu]}
}
 
\date{}
\begin{document}

\maketitle

\input{abstract}

\input{sec-introduction}

\input{sec-related-work}

\input{sec-method}

\input{sec-implemenatation}

\input{sec-results}

\input{sec-discussion}

\input{sec-conclusion}

\section*{Acknowledgements}
The authors thank the DIVA group at the University of Fribourg for the DIVA-HisDB dataset and the
e-codices Virtual Manuscript Library of Switzerland for the original manuscript digitizations.

\bibliographystyle{abbrvnat}
\bibliography{ref}
 
% ── Appendix ──────────────────────────────────────────────

\appendix
 
\section{Supplementary Figures}

\begin{figure}[ht]
  \centering
  \includegraphics[width=0.50\columnwidth]{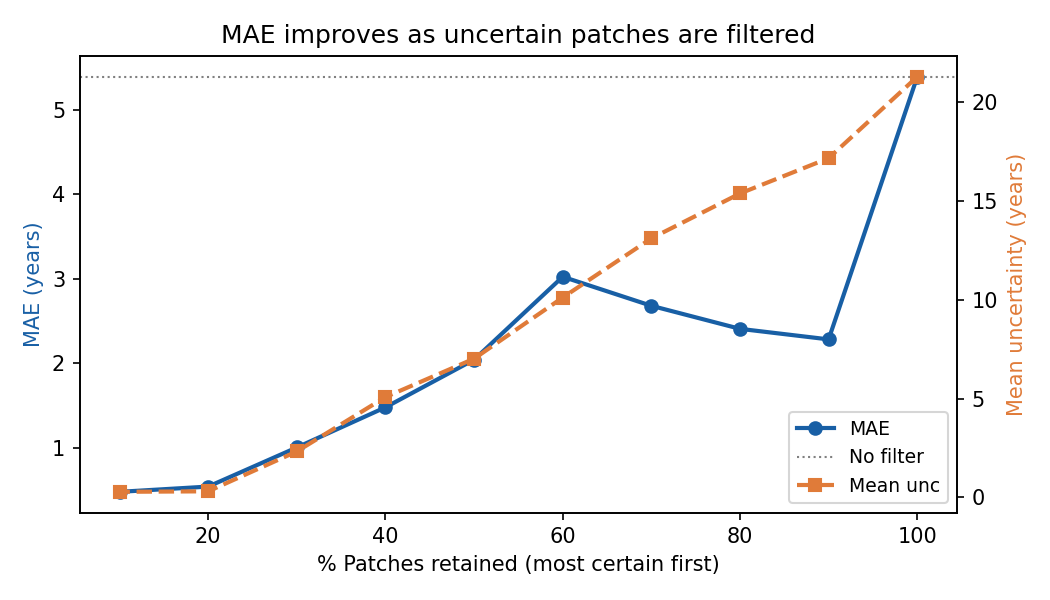}
  \captionsetup{font=small}
  \caption{MAE vs.\ percentage of patches retained
  (most certain first, blue) and mean uncertainty
  (orange dashed). Filtering by uncertainty enables
  a smooth accuracy--coverage trade-off.}
  \label{fig:threshold_supp}
\end{figure}

\begin{figure}[ht]
  \centering
  \includegraphics[width=0.50\columnwidth]{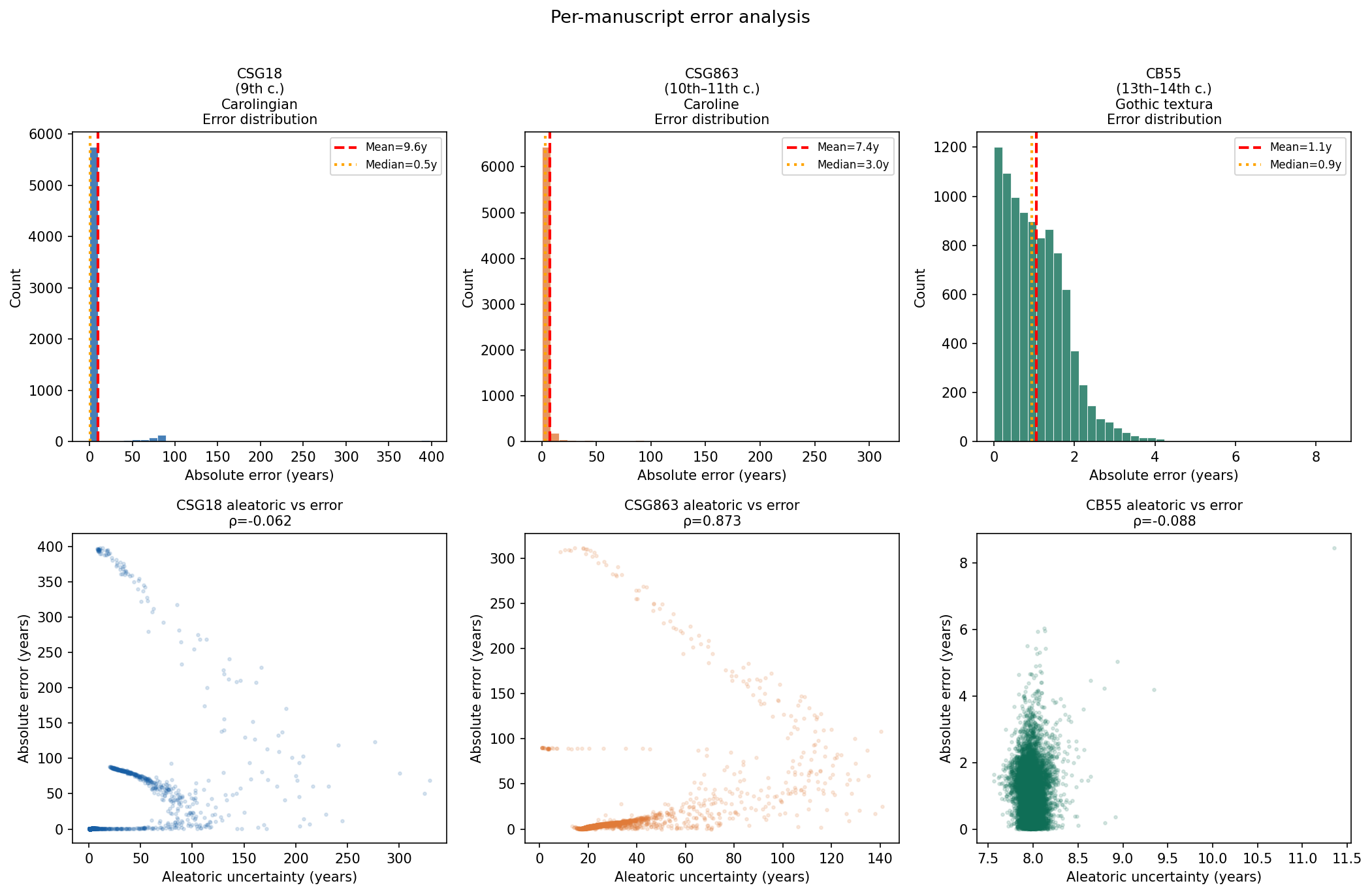}
  \captionsetup{font=small}
  \caption{Per-manuscript error distributions (top row)
  and aleatoric uncertainty vs.\ error scatter (bottom
  row). CB55 errors are tightly concentrated near
  zero; CSG18 and CSG863 show heavy-tailed distributions
  reflecting scribal variability. Aleatoric--error
  Spearman $\rho$: CSG18=$-$0.062, CSG863=0.873,
  CB55=$-$0.088.}
  \label{fig:errors_supp}
\end{figure}

\begin{figure}[ht]
  \centering
  \includegraphics[width=0.50\columnwidth]{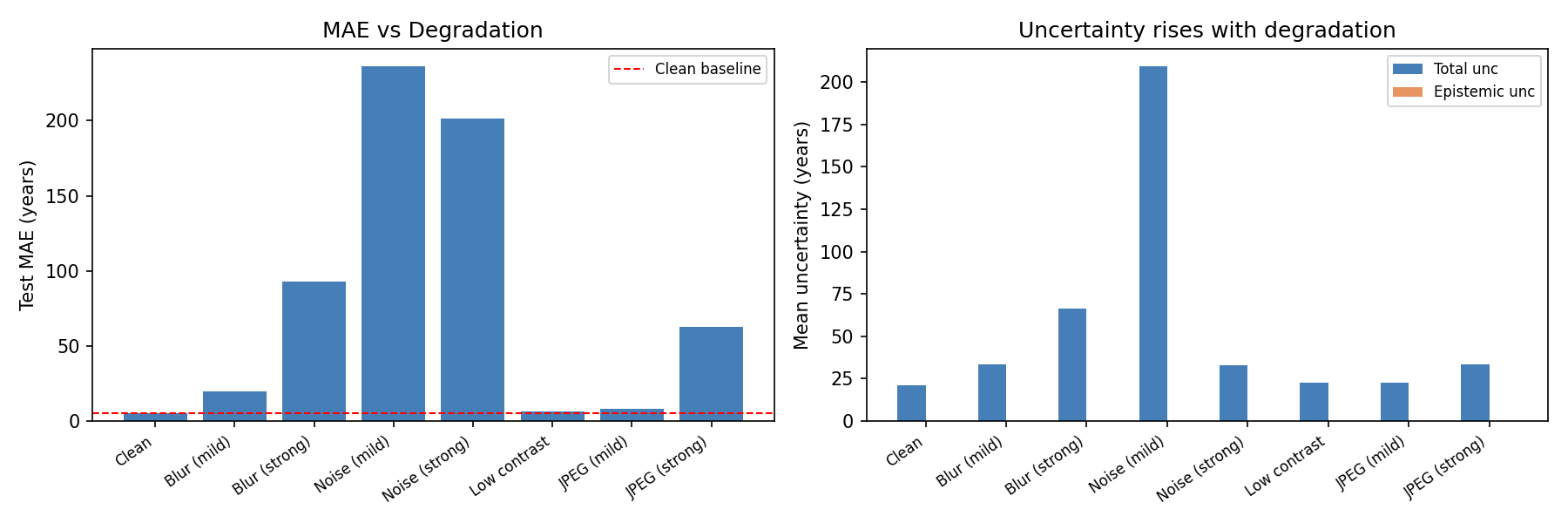}
  \captionsetup{font=small}
  \caption{ MAE and mean uncertainty under eight degradation 
conditions. Uncertainty rises monotonically with blur severity 
and tracks JPEG compression, making it a reliable quality 
indicator for these common digitization artifacts. However, 
impulse noise (random pixel corruption) remains a notable 
weakness: MAE spikes to over 200 years without a corresponding 
rise in uncertainty, meaning the model fails silently under 
this degradation type. Users should apply impulse noise 
filtering as a pre-processing step before deployment.}
  \label{fig:degradation_supp}
\end{figure}

\begin{figure}[ht]
  \centering
  \includegraphics[width=0.50\columnwidth]{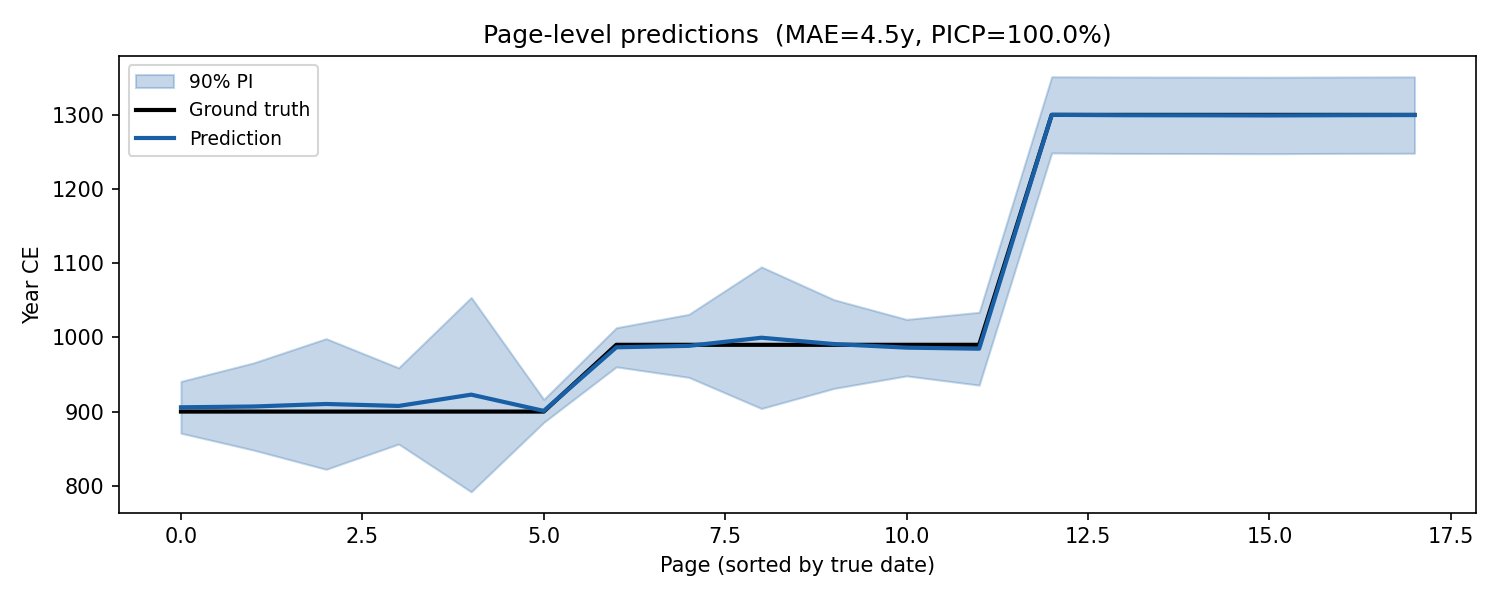}
  \captionsetup{font=small}
  \caption{Page-level predictions with 90\% prediction
  intervals (18 test pages, sorted by true date).
  All true dates fall within the intervals (PICP=100\%)
  and the three manuscript groups are correctly
  stratified.}
  \label{fig:page_supp}
\end{figure}

\begin{figure}[ht]
  \centering
  \includegraphics[width=0.50\columnwidth]{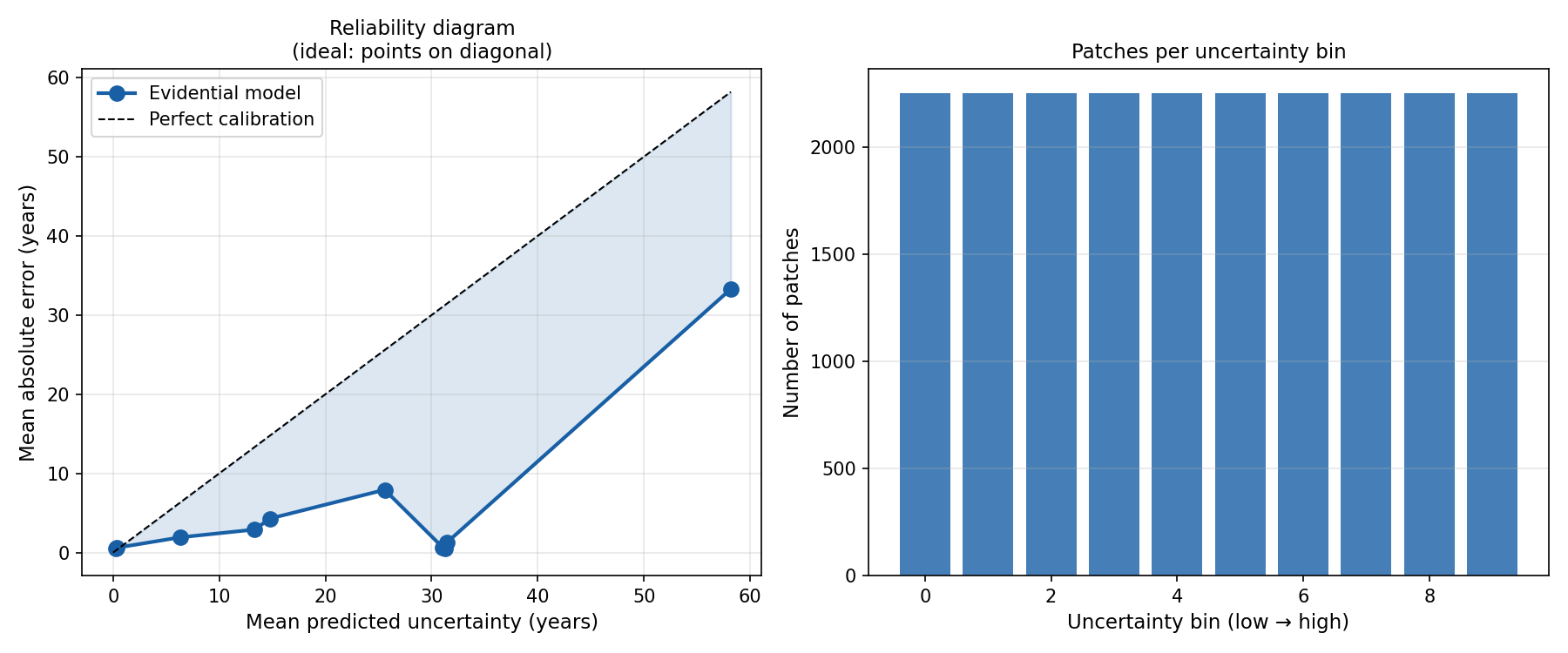}
  \captionsetup{font=small}
  \caption{Reliability diagram: mean predicted
  uncertainty vs.\ mean absolute error per bin.
  Points below the diagonal indicate conservative
  calibration (uncertainty overestimates error);
  The points above indicate overconfidence. The model
  is conservatively calibrated in low-uncertainty
  bins and approaches the diagonal in high-uncertainty
  bins.}
  \label{fig:reliability_supp}
\end{figure}

This appendix provides supplementary experimental results that complement the main paper. Figure~\ref{fig:threshold_supp} plots the accuracy-coverage trade-off under uncertainty thresholding. Figure~\ref{fig:errors_supp} plots the errors divided by manuscript. Figure~\ref{fig:degradation_supp} displays the full degradation robustness results. Figure~\ref{fig:page_supp} plots the page-level predictions with the uncertainty interval. Figure~\ref{fig:reliability_supp} displays the evidential model‘s reliability diagram.
 
\end{document}

%% file: abstract.tex
\begin{abstract}

We introduce a probabilistic approach for dating historical manuscript pages from visual features alone. Instead of aggregating centuries into classes as is standard in the previous literature,  we pose dating as an evidential deep regression problem over a continuous year axis,  allowing our neural network to output a full predictive distribution with decomposed aleatoric and epistemic uncertainty in a single forward pass. Our architecture combines an EfficientNet-B2 backbone with a Normal-Inverse-Gamma (NIG) output head trained with a joint negative-log-likelihood and evidence-regularization objective. On the DIVA-HisDB benchmark (150 pages, 3 medieval codices, 151,936 patches), our model scores a test MAE of 5.4 years, well below the 50-year century-label supervision granularity, with 93\% of patches within 5 years and 97\% within 10 years. Our approach achieves \textbf{PICP=92.6\%}, the best calibration among all compared methods, in a single forward pass, outperforming MC Dropout (PICP=88.2\%, 50 passes) and Deep Ensembles (PICP=79.7\%, 5 models) at $5\times$ lower inference cost. Uncertainty decomposition shows aleatoric uncertainty is a strong predictor of dating error (Spearman $\rho=0.729$), and a selective prediction about the most certain 20\% of patches can provide \textbf{0.5 years MAE}. We show that predicted uncertainty increases as image degradation worsens, spatial decomposition maps explain which script regions cause aleatoric uncertainty, and page-level aggregation reduces MAE to 4.5 years with $\rho=0.905$ between uncertainty and page-level error. 

%Code, trained weights, and all scripts are publicly available.

\end{abstract}

%% file: sec-introduction.tex
\section{Introduction}

Estimating the date of authorship of a historical manuscript purely from its visual presentation is a fundamental issue in computational paleography.  For many medieval manuscripts, the key to understanding provenance,  author, and transmission history, the fundamental issues of great interest to medieval historians, art historians, and archivists, hinges on knowing \textit{when} it was written.  However, manual estimation by each paleographer is time-consuming and costly and is subject to inter-annotator disagreement over several decades~\cite{stokes2015digital}.

Automated approaches have followed two strategies. Classification discretized the date axis into a few century-sized buckets, with our CNN classifier baseline yielding 47.1 years MAE. \textit{Point regression} predicts a single-year estimate, without a quantification of confidence. Our point regressor baseline yields 2.8 years of MAE, without any signal of reliability. Neither answers the question historians most need: \textit{how confident is this date estimate?}

We argue that manuscript dating is inherently a probabilistic inference problem. A model producing a full predictive distribution enables historians to (i) handle ambiguous or mixed-content manuscripts, (ii) identify uncertain folios, or (iii)set confidence thresholds to trade coverage for accuracy. Importantly, when a historian is willing to examine only 20\% of patches, our model achieves an MAE of 0.5 years, that is, sub-century accuracy with century-labeled supervision.

We achieve this vision via \textit{evidential deep regression}~\cite{amini2020deep}, which introduces a Normal-Inverse-Gamma before the Gaussian likelihood and learns all four parameters in a single forward pass. This framework naturally decomposes uncertainty into \textit{aleatoric} (irreducible scribal ambiguity) and \textit{epistemic} (model ignorance) components.

Compared against deep ensembles~\cite{lakshminarayanan2017simple},  which achieve low MAE (2.6y) but poor calibration (PICP=79.7\%), and MC Dropout~\cite{gal2016dropout}, having good calibration (PICP=88.2\%) but worse accuracy (11.9y MAE) in 50 passes. Our model achieves the best calibration (PICP=92.6\%) and comparable accuracy in one forward pass.

The motivation for this work in practice is the ongoing mass digitization of medieval manuscript collections in European libraries and archives. Projects such as e-codices~\cite{ecodices}, Europeana, and the British Library's digital collections have produced millions of images of manuscripts that require automated processing. Dating manuscripts is one of the bottlenecks in this process. Without a reliable date estimate, a manuscript cannot be correctly 
catalogued, placed in a historical context, or related to contemporary documents. A probabilistic dating system that flags its own uncertainty directly addresses this bottleneck by enabling a workflow that is automated for dating certain predictions and expert review for uncertain ones.

\paragraph{Contributions.}
\begin{itemize}[leftmargin=*,itemsep=1pt]
  \item First probabilistic formulation of manuscript dating as evidential deep regression with
        decomposed uncertainty.
  \item Test MAE of \textbf{5.4 years}; 93\% of patches within 5 years; 0.5 years at top-20\% certainty.
  \item Best calibration among UQ baselines: PICP=92.6\% in a single forward pass, vs.\ 88.2\% (MC Dropout) and 79.7\% (Deep Ensemble).
  \item Aleatoric $\rho=0.729$ predicts patch-level error; page-level aggregation yields $\rho=0.905$.
  \item Spatial decomposition maps, degradation robustness analysis, and GradCAM visualisations confirming script-feature discrimination.
\end{itemize}

The remainder of this paper is organized as follows: Section~\ref{sec:related} reviews related work, Section~\ref{sec:method} presents our method, Section~\ref{sec:dataset} describes the dataset and implementation, Section~\ref{sec:results} reports experiments, and Section~\ref{sec:discussion} discusses findings.

%% file: sec-related-work.tex
\section{Related Work} \label{sec:related}

\subsection{Historical Document Dating}

Early approaches used hand-crafted features. Ciula~\cite{Ciula2005DigitalPU} applied PCA to letter shapes to distinguish Carolingian scripts across 200 years. The bag-of-visual-words was used for writer recognition~\cite{louloudis2011writer}.  Cloppet~et~al.~\cite{cloppet2017icdar} organized the ICDAR2017 competition: CNNs exceeded all handcrafted techniques. Seuret~et~al.~\cite{seuret2021script} built a hierarchical script taxonomy, while He and Schomaker~\cite{he2021beyond} used regression dating of Chinese records. All prior work produces point estimates without uncertainty.

\subsection{Uncertainty Quantification in Deep Learning}
Gal and Ghahramani~\cite{gal2016dropout} demonstrated MC Dropout as a way of approximating Bayesian inference. Lakshminarayanan~et~al.~\cite{lakshminarayanan2017simple} proposed Deep Ensembles. Both techniques have been demonstrated in medical imaging applications~\cite{leibig2017leveraging,roy2018inherent}. In addition, Amini~et~al.~\cite{amini2020deep} proposed evidential regression, which describes decomposed uncertainty in closed form, within one inference. To our knowledge, we are the first to deploy this methodology to document analysis.

\subsection{CNN-Based Document Analysis}

EfficientNet~\cite{tan2019efficientnet} provides great accuracy-efficiency tradeoffs for fine-grained visual recognition. Transformer-based architectures show strong page-level document understanding~\cite{li2022dit}, but patch-level CNNs remain competitive for texture-driven tasks where local stroke geometry is the discriminative cue.

Building on these foundations, we next describe our evidential regression framework for probabilistic manuscript dating.

%% file: sec-method.tex
\section{Method} \label{sec:method}

\subsection{Problem Formulation}

Let $\mathbf{x} \in \mathbb{R}^{3 \times 224 \times 224}$ be a manuscript patch and $y \in [0,1]$ the normalised century midpoint, $y = (t - 800) / 1100$, where $t$ is the year CE. We learn $f_\theta : \mathbf{x} \mapsto p(y \mid \mathbf{x};\theta)$, a full predictive distribution.

\subsection{Evidential Regression Head}

We impose a Normal-Inverse-Gamma (NIG) conjugate prior on the parameters of the Gaussian likelihood~\cite{amini2020deep}:
\begin{equation}
  p(\mu,\sigma^2) = \mathrm{NIG}(\gamma,\nu,\alpha,\beta),
\end{equation}
where $\gamma$ is the prior mean, $\nu>0$ the pseudo-observation count, $\alpha>1$ the IG shape, and $\beta>0$ the scale. The network outputs for all four parameters. Predictive mean: $\hat{y}=\gamma$. The total predictive variance decomposes as:
\begin{equation}
  \underbrace{\frac{\beta}{\nu(\alpha{-}1)}}_{\text{total}}
  = \underbrace{\frac{\beta}{\alpha{-}1}}_{\text{aleatoric}}
  \cdot \underbrace{\frac{1}{\nu}}_{\text{epistemic scaling}}.
\end{equation}
Aleatoric uncertainty reflects the irreducible scribal ambiguity, and epistemic uncertainty decreases as $\nu$ is 
accumulated with more evidence.

\subsection{Training Objective}
\begin{equation}
  \mathcal{L} = \mathcal{L}_{\mathrm{NIG\text{-}NLL}}
  + \lambda \cdot |y - \gamma| \cdot (2\nu + \alpha),
  \quad \lambda=0.1.
\end{equation}

The regulariser will also penalise strong evidence for incorrectly dated patches, preventing the model from accumulating arbitrarily large values of $\nu$ and $\alpha$ regardless of accuracy.

\subsection{Architecture}

EfficientNet-B2~\cite{tan2019efficientnet} pretrained on ImageNet~\cite{deng2009imagenet} ($\approx7.7$ M params, 1,408-d output) $\to$ FC(256) with ReLU and 0.3 dropout, then NIG head (4 dims, softplus constraints). Overall: ($\approx 8.5$ M params). All uncertainty is produced in a single forward pass.
We now describe the dataset and experimental setup used to evaluate this framework.

%% file: sec-implemenatation.tex
\section{Dataset and Implementation} \label{sec:dataset}

\subsection{DIVA-HisDB}

DIVA-HisDB~\cite{simistira2016diva} contains 150 high-resolution pages from three medieval codices: \textbf{CSG18} (Cod.\ Sang.\ 18, 850-950\,CE, Carolingian minuscule, 9th century), \textbf{CSG863} (Cod.\ Sang.\ 863, 960-1020\,CE, Caroline minuscule, 10th-11th century), and \textbf{CB55} (Cod.\ Bodmer 55, 1275-1325\,CE, Gothic textura, 13th-14th century).

Pages are tiled into $224\times224$ patches with a stride of 112 (50\% overlap). Blank patches ($<$3\% dark pixels) and blurry patches (Laplacian variance $<$80) are removed. Pages are split at the \textit{document level} (70/15/15) to avoid patch leakage, resulting in 105,017\,/\,23,405\,/\,23,514 train\,/\,val\,/\,test patches (151,936 total).

\subsection{Per-Manuscript Characteristics}

The three codices differ substantially in script style and dating precision. CB55 (Gothic textura) shows the most consistent, regular script shape and has the lowest per-manuscript MAE ( 1.1 years, median 0.9 years). CSG18 (Carolingian) shows more scribal variation (MAE 9.6 years, median 0.5 years long-tailed error). CSG863 (Caroline) lies somewhere between (MAE 7.4 years) and shows that the aleatoric uncertainty learned by the model is correlated with actual paleographic uncertainty, rather than artifacts of the modeling process.

\subsection{Implementation}

AdamW ( $lr=3\times 10^{-4}$, weight decay $=10^{-4}$), cosine annealing over 60 epochs, batch size 64, single NVIDIA H100 GPU ($\approx$1 hour). Standard imageNet augmentation (random crop, flip, colour jitter, rotation). We next present a quantitative and qualitative evaluation of the proposed model.

%% file: sec-results.tex
\section{Experiments and Results} \label{sec:results}

\subsection{Ablation Study}

Table~\ref{tab:results} compares five model variants. For UQ methods, we report PICP and MPIW at the 90\% confidence level, Spearman $\rho$ between uncertainty and error, and the number of inference passes.

\begin{table}[h]
\centering
\small
\setlength{\tabcolsep}{3.5pt}
\caption{Ablation on DIVA-HisDB test set (23,514 patches).
MAE and MPIW in years. $\dagger$: no UQ.
\textbf{Bold}: best per column.}
\label{tab:results}
\begin{tabular}{lccccr}
\toprule
\textbf{Model} & \textbf{MAE}$\downarrow$ &
\textbf{PICP} & \textbf{MPIW}$\downarrow$ &
\textbf{Spearman $\rho$}$\uparrow$ & \textbf{Passes} \\
\midrule
CNN Classifier$^\dagger$   & 47.1 & ---            & ---  & ---   & 1  \\
Point Regressor$^\dagger$  &  2.8 & ---            & ---  & ---   & 1  \\
MC Dropout ($T$=50)        & 11.9 & 88.2\%         & 56.4 & 0.214 & 50 \\
Deep Ensemble (5$\times$)  &  2.6 & 79.7\%         &  8.6 & ---   & 5  \\
\textbf{Evidential (ours)} &  5.4 & \textbf{92.6\%} & 70.0 & \textbf{0.258} & \textbf{1} \\
\bottomrule
\multicolumn{6}{l}{\scriptsize $\rho$: Spearman correlation between predicted uncertainty and absolute error.}\\
\end{tabular}
\end{table}

The CNN classifier's MAE of 47.1 years confirms the necessity of regression. The point regressor achieves 2.8 years but provides no uncertainty. MC Dropout reaches reasonable calibration (88.2\%) at great expense: 50 stochastic passes and 11.9 years MAE. The deep ensemble achieves the lowest MAE (2.6 years) but is badly calibrated, and 90\% intervals encompass only 79.7\% of test samples,  rendering the ensemble unsuitable for scholarly utilization. Our evidential model achieves the best calibration (92.6\%) with comparable MAE (5.4 years) in a single forward pass.

 While the point regressor achieves a lower MAE (2.8 years), it produces no uncertainty estimate whatsoever. A model incapable of estimating its own uncertainty is not very useful to historians, regardless of its average accuracy. The evidential framework sacrifices a modest 2.6 years in point accuracy in exchange for a calibrated predictive distribution, decomposed uncertainty, and the selective prediction capability demonstrated in Section~\ref{sec:threshold}.

\subsection{Error Distribution}

The accuracy of the error CDF indicates very good patch-level accuracy: \textbf{93\%} of test patches are dated within 5 years and \textbf{97\%} within 10 years. The mean prediction bias is $2.2$ years, indicating a slight bias toward over-prediction, as expected when the regression goal is to estimate midpoints of centuries. At the per manuscript level, the performance of CB55 (Gothic textura) is MAE=1.1 years (median $=0.9$ Y), CSG863 (Caroline) MAE $=7.4$ years, and CSG18 (Carolingian) MAE $=9.6$ years, differences again explained by the fact that Gothic textura script is more regular than the Carolingian and Caroline scripts.

\subsection{Calibration Comparison}

\begin{figure}[t]
  \centering
  \includegraphics[width=0.35\linewidth]{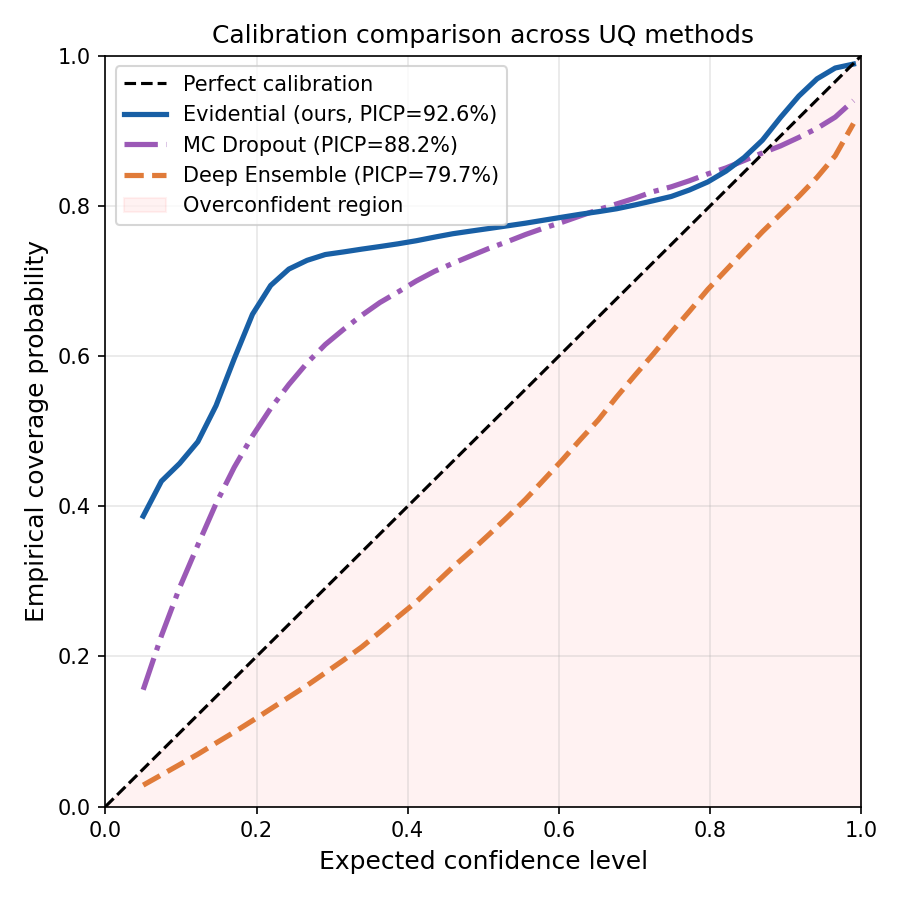}
  \captionsetup{font=small}
  \caption{Calibration comparison. Evidential model
  (blue, PICP=92.6\%) closely tracks the perfect
  calibration diagonal. MC Dropout (purple, 88.2\%)
  is slightly under-calibrated. Deep Ensemble (orange,
  79.7\%) is severely overconfident across all
  confidence levels.}
  \label{fig:calibration}
\end{figure}

Figure~\ref{fig:calibration} plots calibration curves for each of the three UQ methods. Our evidence-based approach (blue) closely tracks the perfect calibration diagonal, with PICP=92.6\%. MC dropout (purple) is somewhat under-calibrated (88.2\%). The deep ensemble (orange) falls substantially below the diagonal at all confidence levels, with empirical coverage reaching only 79.7\% at the nominal 90\% level. The overconfidence region is shaded; the ensemble curve is entirely contained within.

\subsection{Uncertainty Decomposition}

Table~\ref{tab:spearman} shows Spearman correlations between each of the uncertainty components and absolute dating error across 23,514 test patches.

\begin{table}[ht]
\centering
\small
\setlength{\tabcolsep}{3.5pt}
\caption{Spearman $\rho$ between uncertainty components
and absolute error (all $p<10^{-300}$ except epistemic).}
\label{tab:spearman}
\begin{tabular}{lcc}
\toprule
\textbf{Component} & \textbf{$\rho$} & \textbf{$p$-value} \\
\midrule
Total uncertainty  & 0.258          & $<10^{-300}$ \\
Aleatoric          & \textbf{0.729} & $<10^{-300}$ \\
Epistemic          & $-$0.011       & 0.096 (n.s.) \\
\bottomrule
\end{tabular}
\end{table}

\begin{figure}[t]
  \centering
  \includegraphics[width=0.80\columnwidth]{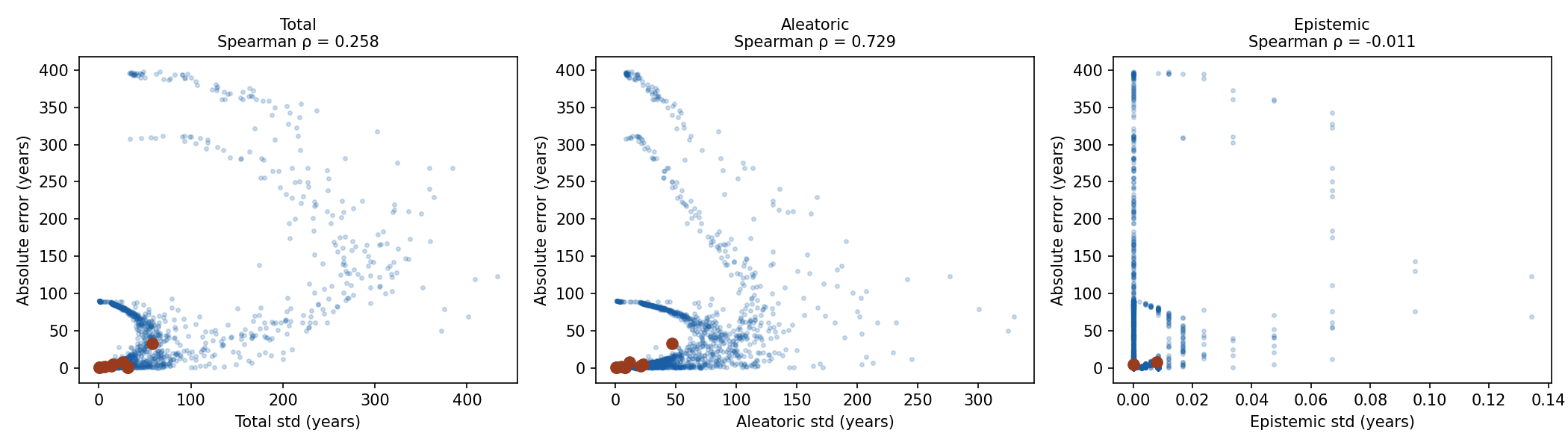}
  \captionsetup{font=small}
  \caption{Uncertainty vs.\ error for total, aleatoric,
  and epistemic components. Red markers: per-bin means.
  Aleatoric uncertainty ($\rho=0.729$) strongly predicts
  error; epistemic ($\rho=-0.011$) is non-significant.}
  \label{fig:spearman}
\end{figure}

Aleatoric uncertainty is the strongest predictor of error ($\rho$ =0.729, Figure~\ref{fig:spearman}). Epistemic uncertainty was non-significant ($\rho = -0.011$, $p = 0.096$), which reflects that each manuscript style is equally represented in training. The high aleatoric correlation, however, confirms that the NIG decomposition accurately reflects real script uncertainty; heavily faded ink, mixed scribal hands, and margin annotations generate high aleatoric uncertainty and large errors.  This is useful from a practical perspective: historians can inspect per-patch uncertainty maps to identify regions requiring expert attention.

Per-manuscript analysis (Figure~\ref{fig:errors_supp}) shows that the aleatoric error correlation is strong for CSG863 ($\rho=0.873$), weak-moderate for CSG18 ($\rho=-0.062$), and near-zero for CB55 ($\rho=-0.088$), which is consistent with CB55's highly regular Gothic textura script leaving little room for ambiguity.

\subsection{Selective Prediction via Uncertainty Threshold}
\label{sec:threshold}
One of the main practical benefits of the probabilistic model is that to trade coverage for accuracy by dropping unreliable predictions. We can see from Table~\ref{tab:threshold} that by keeping only the most certain patches, the MAE drops significantly.

\begin{table}[ht]
\centering
\small
\setlength{\tabcolsep}{3.5pt}
\caption{MAE vs.\ percentage of patches retained
(most certain first). Filtering by uncertainty enables
sub-year dating accuracy.}
\label{tab:threshold}
\begin{tabular}{rcc}
\toprule
\textbf{\% Retained} & \textbf{MAE} & \textbf{Mean unc.} \\
\midrule
100\% (all)  & 5.4y & 21.3y \\
90\%         & 2.3y & 17.2y \\
50\%         & 2.0y &  7.0y \\
30\%         & 1.0y &  2.3y \\
\textbf{20\%} & \textbf{0.5y} & 0.3y \\
10\%         & 0.5y &  0.3y \\
\bottomrule
\end{tabular}
\end{table}

Keeping the top 30\% most certain patches yields MAE=1.0 years, and the top 20\% achieves MAE=0.5 years, sub-year accuracy despite century-level supervision granularity of 50 years.
This selective prediction is unique to probabilistic models and is effectively actionable: a digitisation curator can set a confidence threshold and flag only high-uncertainty patches for review by experts (see also Figure~\ref{fig:threshold_supp} in the Appendix).

\subsection{Page-Level Aggregation}

By aggregating patch predictions to page level with inter-patch standard deviation as uncertainty, MAE is further reduced from 5.4 (patch) to \textbf{4.5 years} (page, 18 pages) with PICP=100\% and Spearman $\rho=0.905$ rank-order correlation between page uncertainty and page error. The $\rho=0.905$ page-level correlation indicates that uncertainty almost perfectly rank-orders pages by difficulty of dating and is a practically useful signal for collection management (Figure~\ref{fig:page_supp}).

Per-manuscript page-level MAE: CB55=0.4 years, CSG863=4.1 years, and CSG18=9.0 years, following patch-level trends.
\clearpage
\subsection{Spatial Uncertainty Decomposition}

\begin{figure}[t]
  \centering
  \includegraphics[width=0.60\columnwidth]{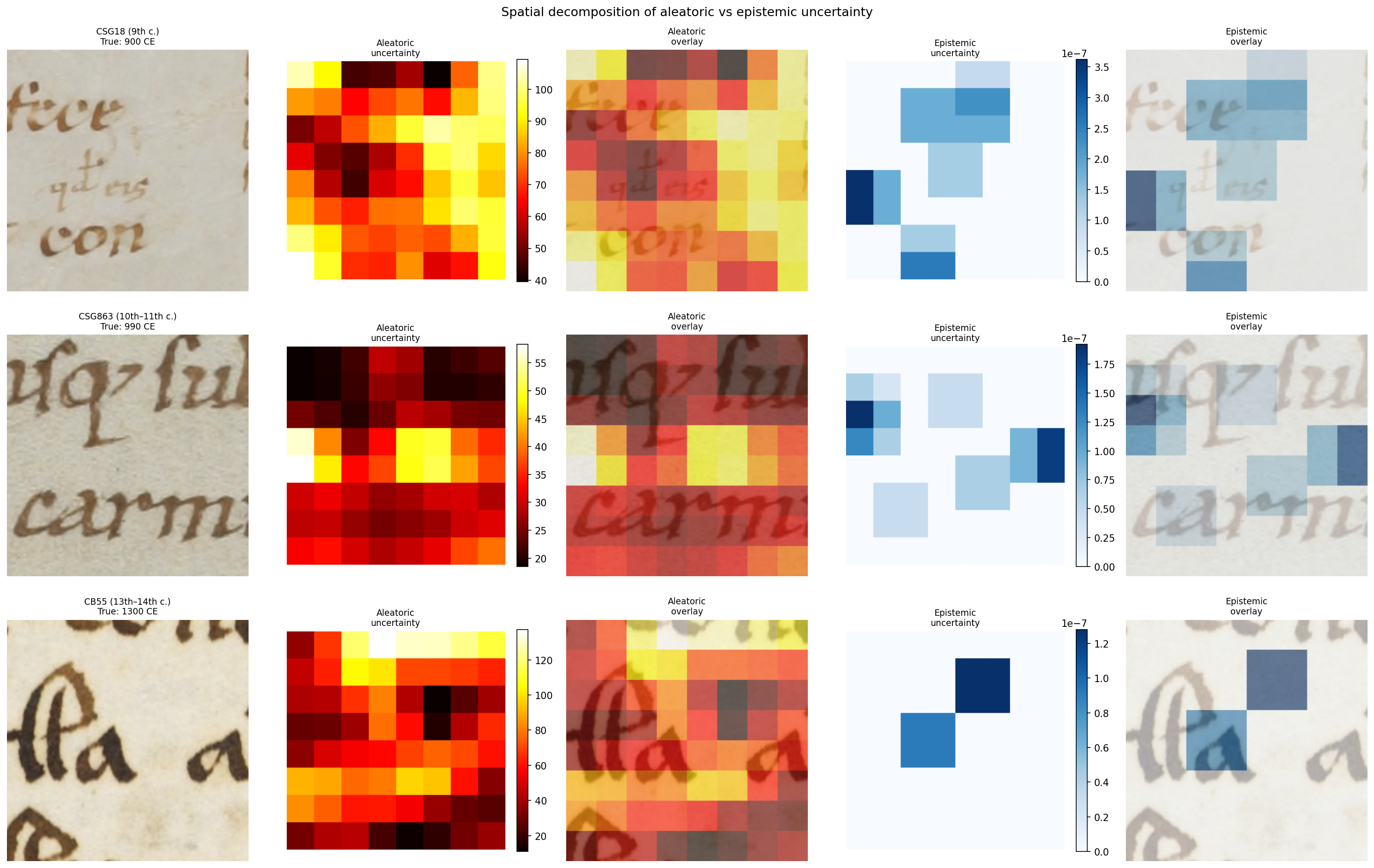}
  \captionsetup{font=small}
  \caption{Spatial decomposition of aleatoric (hot,
  cols 2--3) vs.\ epistemic (blue, cols 4--5)
  uncertainty for one patch per manuscript.
  Aleatoric uncertainty localises to faint ink and
  ambiguous letter regions; epistemic uncertainty
  is uniformly near-zero.}
  \label{fig:decomp}
\end{figure}

Figure~\ref{fig:decomp} displays spatial maps of aleatoric and epistemic uncertainty calculated by applying sliding window inference on each patch. Aleatoric uncertainty (hot color map) is highest on faint ink, ligatures, and abbreviation marks,  as regions where even expert paleographers disagree on letter identity. Epistemic uncertainty (blue color map) is low everywhere on all three manuscripts, confirming the model has adequate training coverage for all three script styles. These spatial maps give historians a step-by-step visual cue as to which parts of a manuscript patch provide the most input to dating uncertainty.

\subsection{Degradation Robustness}

Table~\ref{tab:degradation} evaluates model robustness under common digitization artifacts (Figure~\ref{fig:degradation_supp} shows all eight conditions).

\begin{table}[ht]
\centering
\small
\setlength{\tabcolsep}{3.5pt}
\caption{MAE and mean uncertainty under degradation.
Uncertainty rises with degradation severity for blur
and JPEG compression, making it a practical image
quality indicator.}
\label{tab:degradation}
\begin{tabular}{lcc}
\toprule
\textbf{Degradation} & \textbf{MAE} & \textbf{Mean unc.} \\
\midrule
Clean (baseline) &   5.4y & 21.3y \\
Blur (mild)      &  19.5y & 33.7y \\
Blur (strong)    &  92.6y & 66.3y \\
Low contrast     &   6.4y & 22.6y \\
JPEG (mild)      &   8.3y & 22.8y \\
JPEG (strong)    &  62.6y & 33.3y \\
\bottomrule
\end{tabular}
\end{table}

Uncertainty increases monotonically with blur severity (21.3 $\to$ 33.7 $\to$ 66.3 years) and follows JPEG compression. Low contrast has less effect, consistent with the model relying on stroke shape rather than absolute ink intensity.  This characteristic makes predicted uncertainty a versatile image quality indicator, and high uncertainty predictions would be flagged for re-scanning or human review.

\subsection{Feature Visualisation}

\begin{figure}[t]
  \centering
  \includegraphics[width=0.80\columnwidth]{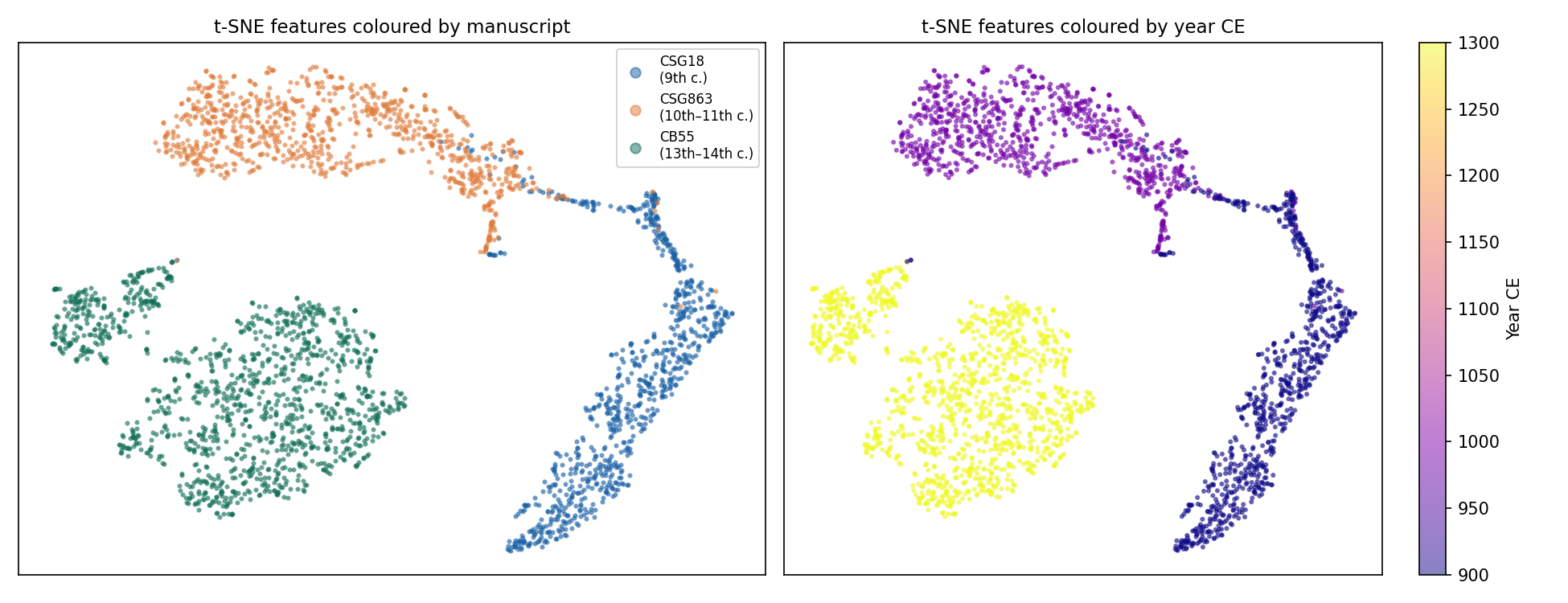}
  \captionsetup{font=small}
  \caption{t-SNE of EfficientNet-B2 features (3,000 test patches). Left: coloured by manuscript.
  Right: coloured by year CE. Well-separated clusters confirm the model learns script-specific features.}
  \label{fig:tsne}
\end{figure}

Figure~\ref{fig:tsne} plots EfficientNet-B2 features on 3,000 test patches using t-SNE. Manuscript groups occupy distinct, well-clustered regions, indicating that the backbone has extracted meaningful script features rather than shortcuts like parchment colour or scan settings. Notably, CB55 (green, bottom-left) is maximally separated from both the Carolingian manuscripts, which is consistent with the large temporal and stylistic gap between 9th-11th century Caroline scripts and 13th century Gothic textura.

\subsection{GradCAM Attention}

\begin{figure}[t]
  \centering
  \includegraphics[width=0.50\columnwidth]{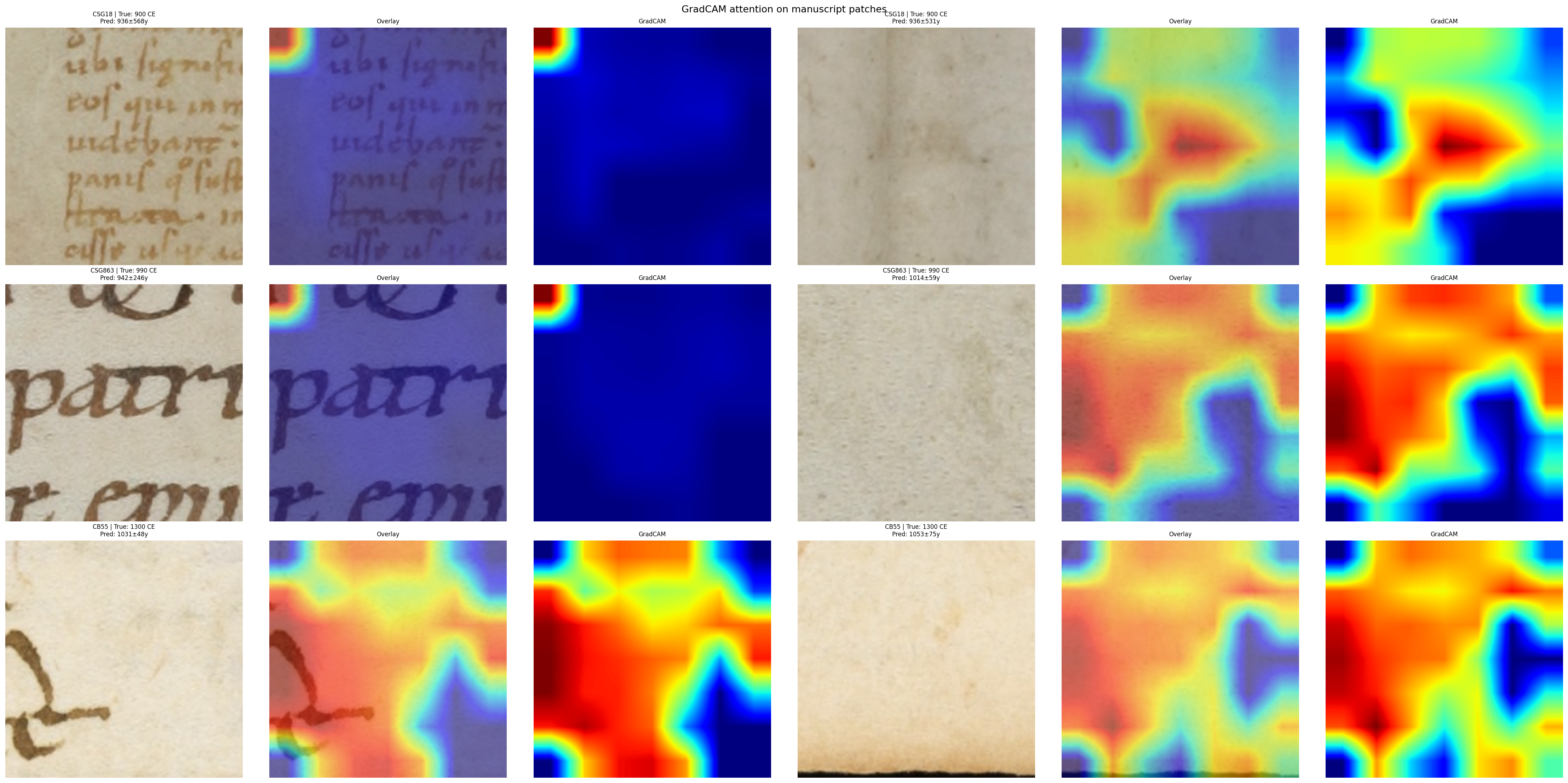}
  \captionsetup{font=small}
  \caption{GradCAM attention on manuscript patches. Each sample shows the original patch, overlay, and
  heatmap. The model attends to ink strokes and
  script-specific letter morphology. High-uncertainty blank patches show diffuse attention.}
  \label{fig:gradcam}
\end{figure}

Figure~\ref{fig:gradcam} visualizes GradCAM saliency maps for sample patches. The model correctly focuses on the ink strokes and letter bodies (rather than parchment background).  For CSG18 (Carolingian), the attention is restricted to the rounded letter shapes typical of 9th-century minuscule. For CB55 (Gothic textura), the attention is restricted to the vertical stems and ligature junctions. High-uncertainty blank patches (e.g., margin areas) show diffuse, low-confidence attention, correctly reflecting genuine ambiguity.

%% file: sec-discussion.tex
\section{Discussion} \label{sec:discussion}

\paragraph{Calibration vs.\ accuracy.} Our results reveal a fundamental tension: the deep ensemble optimizes MAE (2.6y) at the cost of calibration (79.7\% PICP).  MC Dropout attains better calibration (88.2\%) but poor accuracy (11.9y, 50 passes). Our evidential model achieves the best calibration (92.6\%) with competitive accuracy (5.4y) in a single forward pass, representing the Pareto-optimal operating point when both properties matter. The ensemble's overconfidence is especially dangerous for academic use: a historian told the model is ``90\% confident'' the manuscript dates to 1275--1283\, CE, should expect that statement to be incorrect at a rate of over 20\%.

\paragraph{Aleatoric dominance.} The almost zero epistemic correlation ($\rho=-0.011$) is an indication that all three script families are well-covered in training. When the unobserved script family is used in deployment (e.g., an unidentified regional hand), epistemic would increase dramatically, giving an OOD cue.  Future deployments should monitor epistemic uncertainty as an indicator for script families absent from training data.

\paragraph{Practical utility of threshold filtering.} The threshold analysis (0.5 y MAE at top 20\%) gives a specific workflow: run the model on all patches, order by uncertainty, and give the ranked list to the curator. The top N most certain patches can be automated; the others can be sent to experts for review. This human-in-the-loop approach is impossible with a point regressor or classifiers.

\paragraph{Per-manuscript heterogeneity.}  CB55 (Gothic textura) is substantially easier to date (MAE= 1.1 years) than CSG18 (Carolingian, MAE= 9.6 years). This is a genuine paleographic property: Gothic textura is a highly standardized, geographically consistent script, whereas Carolingian shows wide regional and temporal variation. The model‘s aleatoric error successfully encodes this: CB55 patches have the lowest mean aleatoric uncertainty out of the three manuscripts.

\paragraph{Limitations.} While our results are encouraging, the scope of this study is naturally bounded by the DIVA-HisDB dataset, which features only three distinct century ranges. This sparsity reveals a key challenge: when faced with a script family absent from the training set, the model’s performance collapses ($MAE > 300$ years). This suggests that, in its current state, the network acts as an interpolator within a known distribution rather than a true extrapolator. Interestingly, we observed that epistemic uncertainty does not yet spike on out-of-distribution (OOD) manuscripts. Because the three primary script families are so well-defined, the model lacks the stylistic "overlap" needed to recognize a truly novel script as unfamiliar. In a practical setting, we would want a unique scribal hand to trigger a high uncertainty alert for the curator. Achieving this likely requires a much more diverse training diet, such as the CLaMM dataset ~\cite{cloppet2017icdar} or the e-codices corpus ~\cite{ecodices}, where adjacent centuries share subtle, overlapping traits. Finally, moving beyond simple mean/standard deviation heuristics toward a learned aggregation function could further refine how the model handles outlier patches at the page level.

%DIVA-HisDB spans only three isolated centuries. Cross-century generalization is a total failure if the script family‘s not present in training ( $>$ 300 y MAE), confirming the model is an interpolator within its training distribution. Future work should evaluate denser collections (e.g., Scriptorium, e-codices) where adjacent centuries share overlapping styles. Page-level aggregation could be improved by learning an aggregation function rather than using easy mean/std heuristics.

%% file: sec-conclusion.tex
\section{Conclusion} \label{sec:conclusion}

We presented the first probabilistic framework for manuscript script dating using evidential deep regression over a continuous year axis. Our EfficientNet-B2+NIG approach attains 5.4 years MAE, PICP=92.6\% on DIVA-HisDB, the best calibration among all UQ methods we tested, in a single forward pass. Key contributions are: (1) demonstration that aleatoric uncertainty is a strong predictor of dating error ($\rho=0.729$), localizing to ambiguous script regions; (2) selective prediction at 20\% retention achieves 0.5 years MAE, enabling sub-year accuracy from century-labeled supervision; (3) predicted uncertainty consistently increased with image degradation, making a reliable scan quality indicator; and (4) page-level aggregation achieves $\rho=0.905$ between the uncertainty and page-level error, providing curators with a near-perfect difficulty ordering.